\def\year{2021}\relax
\documentclass[letterpaper]{article} % DO NOT CHANGE THIS
\usepackage{aaai21}  % DO NOT CHANGE THIS
\usepackage{times}  % DO NOT CHANGE THIS
\usepackage{helvet} % DO NOT CHANGE THIS
\usepackage{courier}  % DO NOT CHANGE THIS
\usepackage[hyphens]{url}  % DO NOT CHANGE THIS
\usepackage{graphicx} % DO NOT CHANGE THIS
\usepackage{natbib}  % DO NOT CHANGE THIS AND DO NOT ADD ANY OPTIONS TO IT
\usepackage{caption} % DO NOT CHANGE THIS AND DO NOT ADD ANY OPTIONS TO IT
\usepackage{verbatim}

\usepackage{enumerate,booktabs,multirow,color,enumitem,indentfirst,bm,amsmath}
\usepackage[colorlinks,linkcolor=red]{hyperref}
\newcommand*\samethanks[1][\value{footnote}]{\footnotemark[#1]}
\title{AnchorFace: An Anchor-based Facial Landmark Detector Across Large Poses }
\author {
        Zixuan Xu\textsuperscript{\rm 1}\thanks{\textbf{These authors contributed equally}},
        Banghuai Li\textsuperscript{\rm 2}\samethanks[1],
        Miao Geng\textsuperscript{\rm 3},
        Ye Yuan\textsuperscript{\rm 2}\\
}
\affiliations {
    \textsuperscript{\rm 1} School of Electronics Engineering and Computer Science, Peking University \\
    \textsuperscript{\rm 2} Megvii Research \\
    \textsuperscript{\rm 3} Beihang University \\
    zixuanxu@pku.edu.cn, \{libanghuai,yuanye\}@megvii.com, geng\_m@buaa.edu.cn
}

\begin{document}

\maketitle

\begin{abstract}
Facial landmark localization aims to detect the predefined points of human faces, and the topic has been rapidly improved with the recent development of neural network based methods.
However, it remains a challenging task when dealing with faces in unconstrained scenarios, especially with large pose variations. 
In this paper, we target the problem of facial landmark localization across large poses and address this task based on a split-and-aggregate strategy. 
To split the search space, we propose a set of anchor templates as references for regression, which well addresses the large variations of face poses. 
Based on the prediction of each anchor template, we propose to aggregate the results, which can reduce the landmark uncertainty due to the large poses. 
Overall, our proposed approach, named AnchorFace, obtains state-of-the-art results with extremely efficient inference speed on four challenging benchmarks, i.e. AFLW, 300W, Menpo, and WFLW dataset. Code will be available at \url{https://github.com/nothingelse92/AnchorFace}.
\end{abstract}

\section{Introduction}
Facial landmark localization, or face alignment, refers to detect a set of predefined landmarks on the human face. It is a fundamental step for many facial related applications, e.g. face verification/recognition, expression recognition, and facial attribute analysis.

With the recent development of convolutional neural network based methods~\cite{ShuffleNet,betarcnn}, the performance for facial landmark localization in constrained scenarios has been greatly improved~\cite{SAN,ODN,LAB}. However, unconstrained scenarios, for example, faces with large pose, still limit the wide application of the existing landmark algorithms. In this paper, we target to address the problem of facial landmark localization across large poses.

\begin{figure}
\centering
\includegraphics[width=9cm]{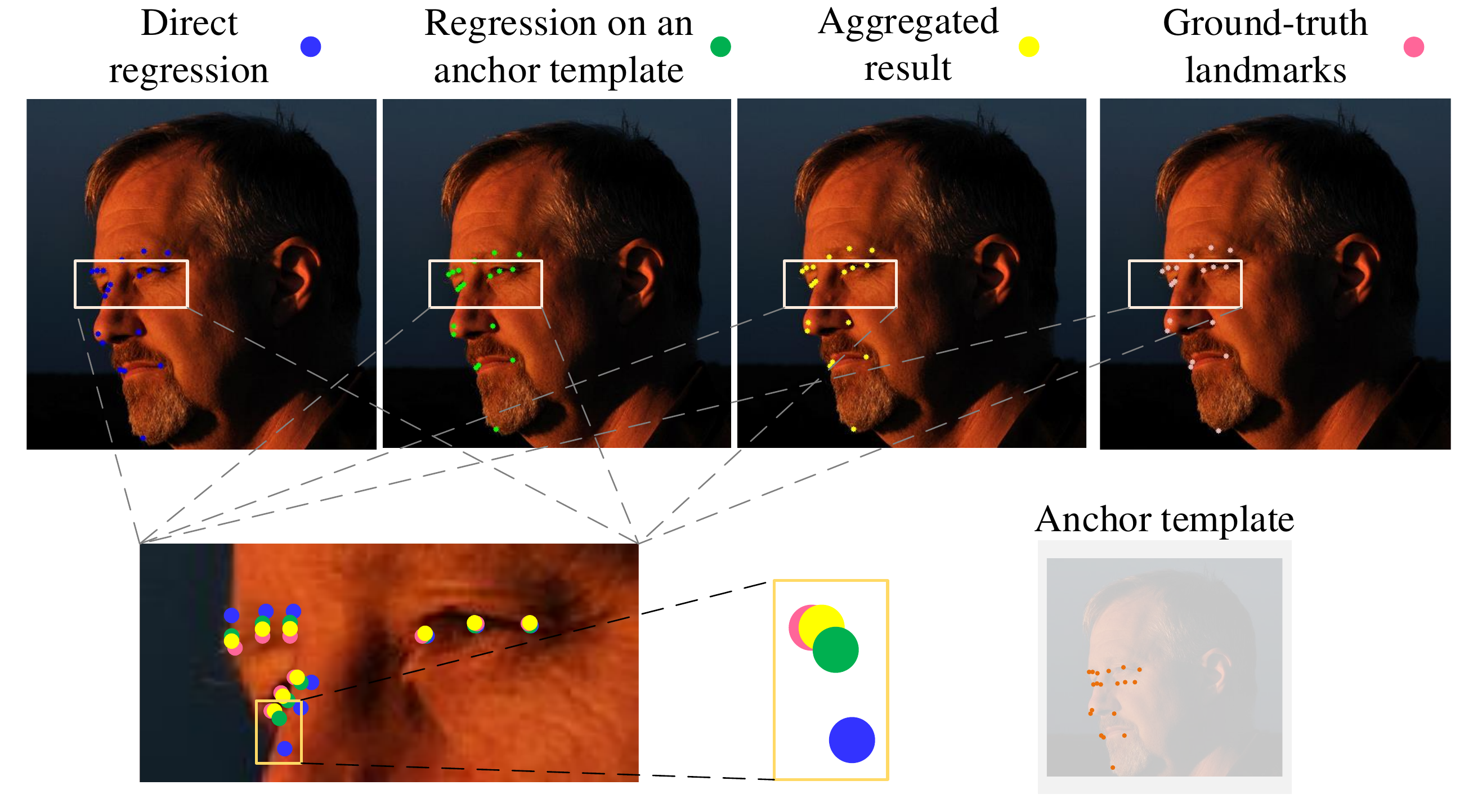}
\caption{A comparison between direct regression and anchor-based regression (AnchorFace). Our AnchorFace includes two steps. The first step is to introduce the anchor templates and regress the offsets based on each anchor template (Second Column). The second step is to aggregate the prediction results from multiple anchor templates (Third Column)}
\label{motivation}
\vspace{0cm}
\end{figure}

There are two challenges for facial landmark detection across large poses. On one hand, faces with large poses will significantly increase the difficulty for landmark localization due to the large variations among different poses. As shown in Fig.~\ref{motivation}, directly regressing the point coordinates may not be able to localize every landmark point precisely. On the other hand, there usually exists a large probability of uncertainty due to the self-occlusion and noisy annotations. For example, occlusion will usually lead to invisible landmarks, which will increase the uncertainty for the landmark prediction. Besides, the faces with a large pose will also cause difficulty during the data annotation process. 

\begin{figure*}[h]
\centering
\includegraphics[width=17cm]{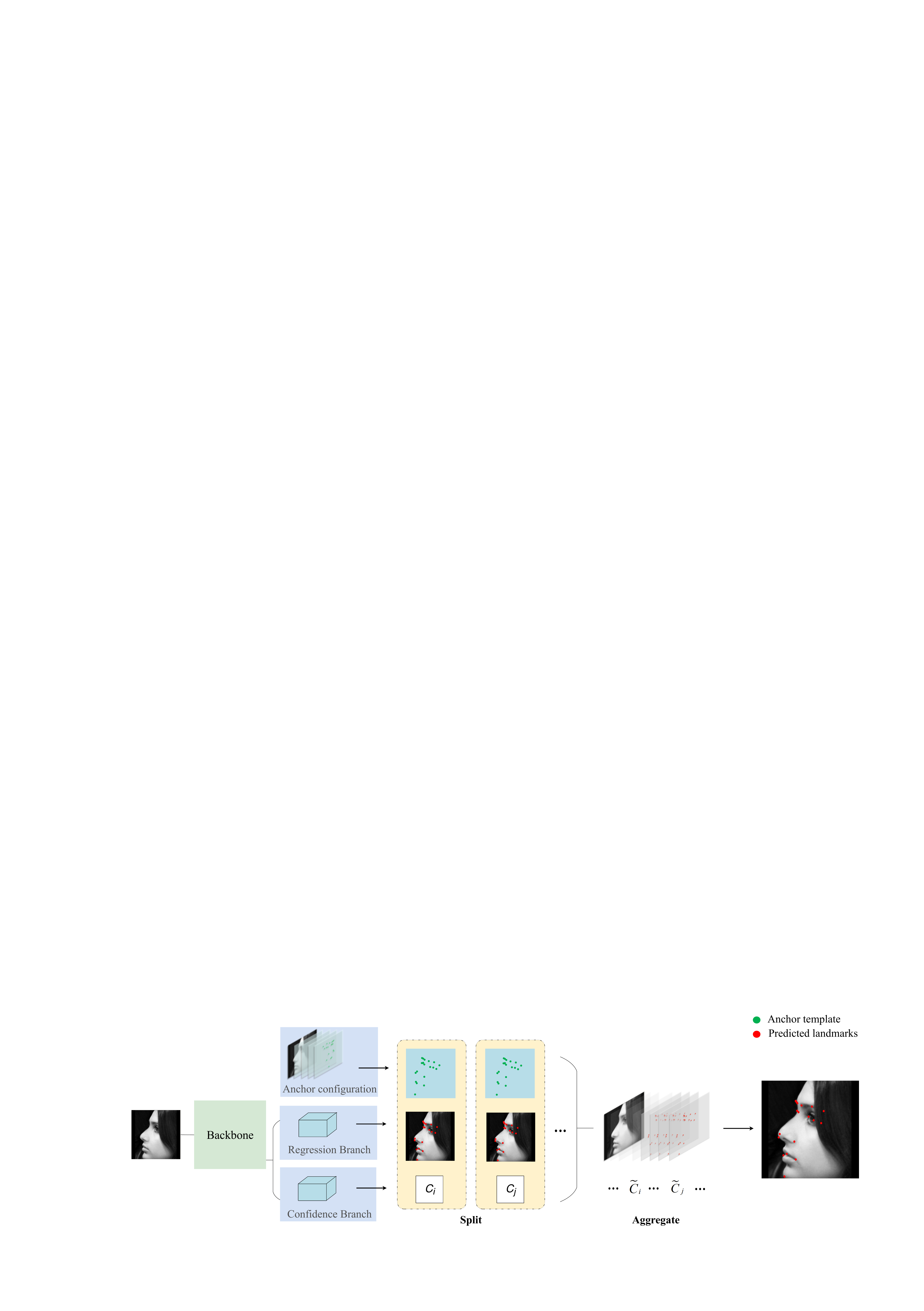}
\caption{The pipeline of our proposed AnchorFace landmark detector. AnchorFace is based on a split-and-aggregate strategy, which consists of the backbone and two functional branches: the offset regression branch and the confidence branch. In the split step, we predict the landmark position based on each anchor template. During aggregate step, the predictions of multiple anchor templates are averaged by weighted confidence}
\label{flow}
% \vspace{-1cm}
\end{figure*}

To address the above two challenges, we propose a novel pipeline for facial landmark localization based on an anchor-based design. The new pipeline includes two steps: split and aggregate. An overview of our pipeline can be found in Fig.~\ref{flow}. To deal with the first challenge with large pose variations, we adopt the divide-and-conquer way following an anchor-based design. We propose to use the anchor templates to split the search space, and each anchor will serve as a reference for regression. This can significantly reduce the pose variations for each anchor. To address the second issue with pose uncertainty, we propose to aggregate each anchor result weighted by the predicted confidence. 

In summary, we propose AnchorFace to implement the split-and-aggregate strategy. There are three contributions in our paper. 
\begin{itemize}
    \item We propose a novel pipeline with a split-and-aggregate strategy which can well address the challenges for face alignment across large poses.
    \item To implement the split-and-aggregate strategy, we introduce the anchor design into the facial landmark problem, which can simplify the search space for each anchor template and meanwhile improve the robustness for landmark uncertainty. 
    \item Our proposed AnchorFace can achieve promising results on four challenging benchmarks with a realtime inference speed of $\sim$45 FPS\footnote{The computational speed of  $\sim$45 FPS is calculated on one Nvidia Titan Xp GPU with batchsize 1. }.
\end{itemize}

\section{Related Work}
\textbf{Facial Landmark Localization.}
In the literature of facial landmark localization, a number of achievements have been developed including the classic ASMs~\cite{ASM}, AAMs~\cite{AAM,AIA}, CLMs~\cite{CLM,CLM2}, and Cascaded Regression Models~\cite{ESR,setting,AFLW-setting}. Nowadays, more and more deep learning-based methods have been applied in this area. These deep learning based methods could be divided into two categories, i.e. coordinate regression methods and heatmap regression methods.

Coordinate regression methods directly map the discriminative features to the target landmark coordinates. The earliest work can be dated to~\cite{Sun_2013_CVPR}. Sun~et al.~\cite{Sun_2013_CVPR} used a three-level cascade CNN to do facial landmark localization in a coarse-to-fine manner, and achieved promising localization accuracy. MDM~\cite{MDM} was the first to apply a recurrent convolutional network model for facial landmark localization in an end-to-end manner. Zhang~et al.~\cite{Zhang_2016_CVPR} utilized a multi-task learning framework to optimize facial landmark localization and correlated facial attributes analysis simultaneously. Recently, Wingloss~\cite{Wing} was proposed as a new loss function for landmark localization, which can obtain robust performance against widely used $L2$ loss.

Heatmap regression methods generate a probability heatmap for each landmark, respectively. Benefit from FCN~\cite{FCN} and Hourglass~\cite{Hourglass}, heatmap regression methods have been successfully applied to landmark localization problems and have achieved state-of-the-art performance. JMFA~\cite{Deng2017JointMF} achieved high localization accuracy with a stacked hourglass network~\cite{Hourglass} for multi-view facial landmark localization in the Menpo~\cite{Menpo-dataset} competition.
Yang~et al.~\cite{Yang_2017_CVPR_Workshops} adopted a supervised face transformation to normalize the faces, then employed an Hourglass network to regress it. Recently, LAB~\cite{LAB} proposed to use additional boundary lines as the geometric structure of a face image to help facial landmark localization. 
% Knowledge distillation also has been applied to facial landmark localization for the lightweight network~\cite{MobileFAN}.

\begin{figure*}[h]
\centering
\includegraphics[width=17cm]{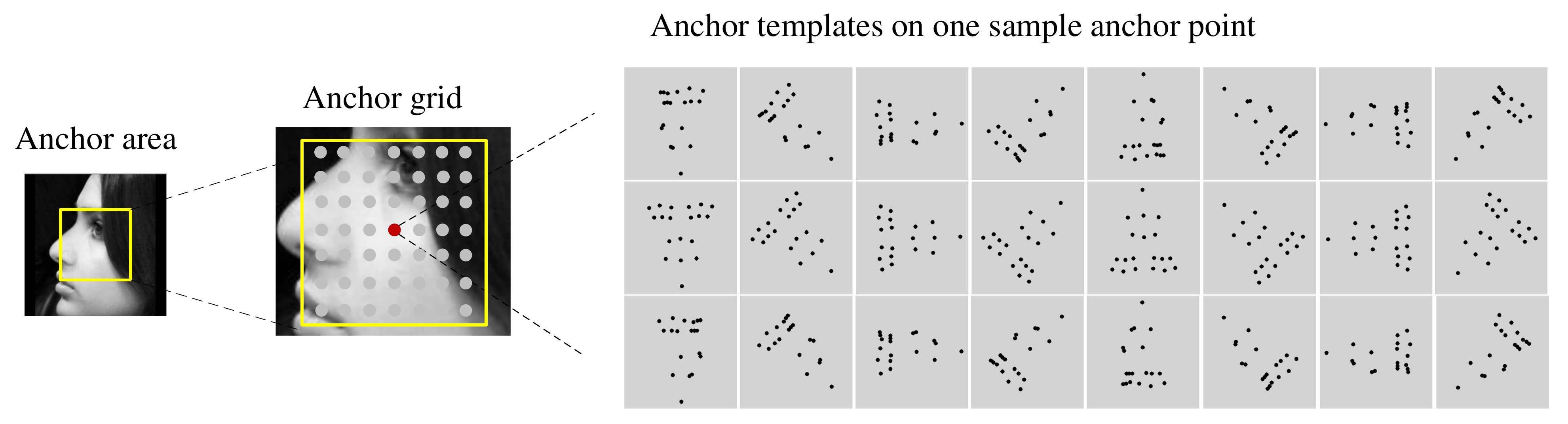}
\caption{An illustration of our anchor configuration. Anchor area is a region centered at the image center with a spatial neighborhood. Based on the anchor area, we setup a grid of anchor points, where each anchor point contains a set of anchor templates to model various pose variations}
\label{templates}
% \vspace{-0.4cm}
\end{figure*}

\textbf{Faces with Large Pose.} 
Large pose is a challenging task for facial landmark localization, and different strategies have been proposed to address the difficulty. Multi-view framework and 3D model are two popular ways. Multi-view framework uses different landmark configurations for different views. For example, TSPM~\cite{TSPM} and CDM~\cite{CDM} employ DPM-like~\cite{DPM} method to align faces with different shape models, and choose the highest possibility model as the final result. However, multi-view methods have to cover each view, making it impractical in the wild. 3D face models have been widely used in recent years, which fit a 3D morphable model~(3DMM)~\cite{3DMM} by minimizing the difference between face image and model appearance. Lost face information can be recovered to localize the invisible facial landmarks~\cite{3D3,3D4,3DDFA,FAR}. However, 3D face models are limited by their own database and the iterative label generation method. Besides, researchers have applied multi-task learning to address the difficulties resulting from pose variations. Other facial analysis tasks, such as pose estimation or facial attributes analysis, can be jointly trained with facial landmark localization~\cite{Multitask2,Multitask1}. With joint training, multi-task learning can boost the performance of each subtask. 
The facial landmark localization task can achieve robust performance. 
But the multi-task framework is not specially designed for landmark localization, it contains much redundant information and contributes to large models.
%: a Basic Landmark Prediction Stage and a Whole Landmark Regression Stage.

In this paper, we propose an anchor-based model for facial landmark localization. Different from~ \cite{A2J}, which utilized anchor points to predict the positions of a human 3D pose, our approach introduces a split-and-aggregate pipeline for the facial landmark localization. Anchor is utilized as a reference for regression in our approach. Overall, our model requires neither cascaded networks nor large backbones, leading to a great reduction in model parameters and computation complexity, while still achieving comparable or even better accuracy.

\section{Proposed Method}

In this paper, we propose a new split-and-aggregate strategy for facial landmark detector across large poses. An overview of our pipeline can be found in Fig.~\ref{flow}. To implement the split-and-aggregate strategy, we introduce the anchor-based design, and our approach is named AnchorFace. In the following section, we will discuss the \textbf{split} and \textbf{aggregate} steps separately, followed by the details on the network training. 

%Face is more like a rigid pattern due to its relatively stable geometric structure. Therefore, we propose an anchor-based facial landmark detector, which uses multiple anchor faces centering anchor points as references for landmark localization. The main pipeline is shown in Fig.~\ref{flow}, and the symbol definitions are listed in Table~\ref{definition}. The network consists of two functional branches: the confidence branch and the regression branch after the backbone. 

\subsection{Split Step}
Due to the large pose variations among different poses, it is a challenging problem to directly regress the facial landmarks while maintaining high localization precision. In this paper, we propose to utilize the divide-and-conquer way to address the issue from large pose variations. More specifically, we propose to employ the anchor templates as regression references to split the search space. Different from the traditional methods which regress the landmarks with a uniform facial landmark detector, we propose to regress the offsets base on a set of anchor templates.

\subsubsection{Anchor Configuration.}\label{sec:anchor_template}
As shown in Fig.~\ref{templates}, there are three hyper-parameters for designing the anchor configuration: anchor area, anchor grid, and anchor templates. 

Anchor area is denoted as the region to set the anchors. It is usually centered at the image center with a spatial neighborhood. The reason to define the anchor area is that the input image is cropped to put the face near the image center. Thus, we select a region near the image center, which is called anchor area, to set up the anchors. Based on the anchor area, we sample a set of anchor points in a grid, e.g. a $7\times7$ grid, as shown in Fig.~\ref{templates}. Each anchor point can be considered as the center of a set of anchor templates. The anchor templates are designed to address the challenges from large pose variations. Intuitively, these anchor templates are used to split the search space for regression and can serve as references for offsets prediction. Therefore, the sampling of anchor templates should be able to cover different variations of large poses and reduce the redundancy for the anchor sets.  

To implement the anchor templates, we present two potential ways. The first one is to hand-design the anchors based on prior knowledge. The second one integrates the proposals generated by the data distribution.

\begin{figure}[h]
    \centering
    \includegraphics[width=8cm]{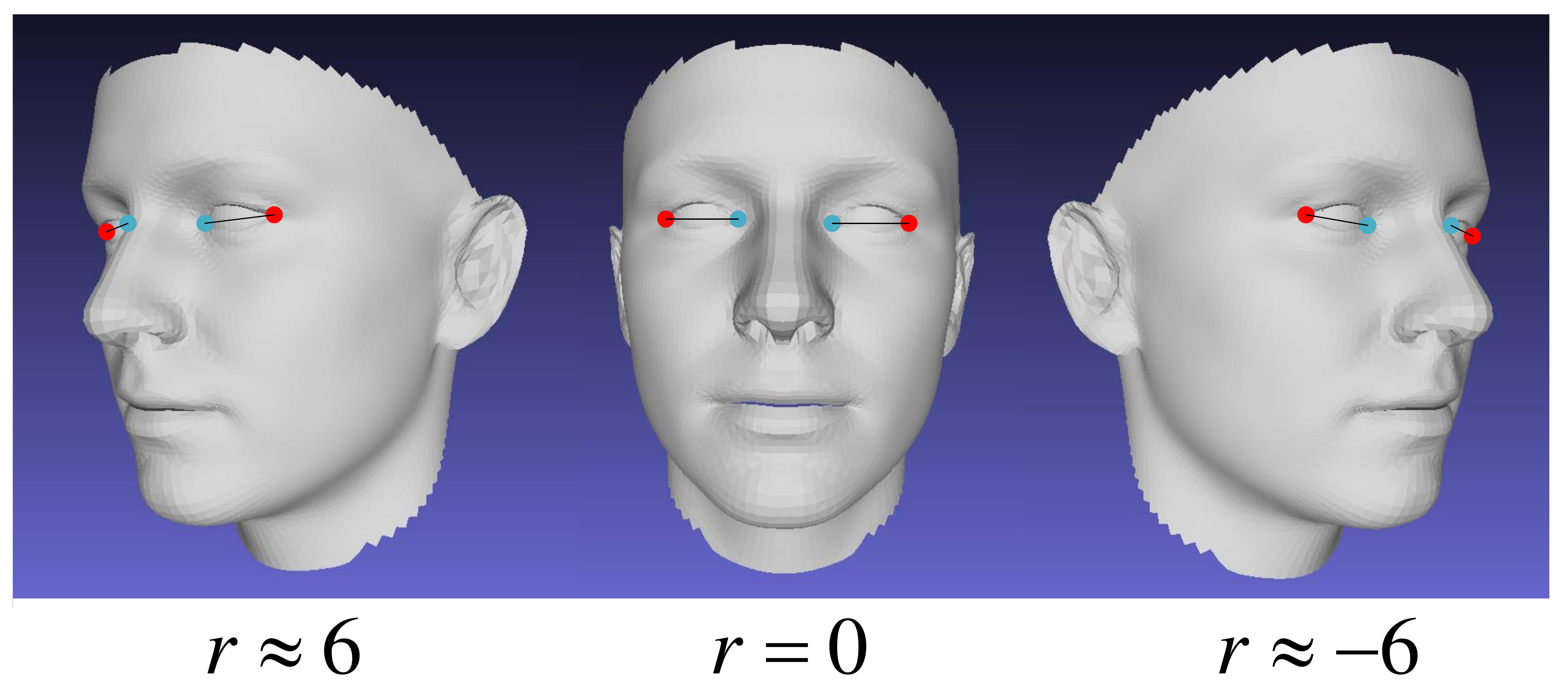}
    \caption{An illustration of the metric $r$ for classifying the faces into three buckets along the yaw direction}
    \label{profile}
\end{figure}

An overview of our hand-designed anchors can be found in Fig.~\ref{templates}. For each anchor point, we explore the 3D pose spaces (yaw, roll, pitch) and design the pose-level anchor set as follows. As unconstrained large-pose faces have large variations on the yaw direction, we first select $N_{yaw}$ base anchors ($N_{yaw}=3$ in our paper representing the anchors for the left, frontal, right faces).  To generate the $N_{yaw}$ base anchors, we utilize a heuristic approach to divide the training faces into three buckets and compute the average face landmarks for each bucket to obtain the anchor proposal. More specifically, we use the ratio of two eyes' width for bucket assignment. We define an indicator to estimate the yaw angle of each training face:
\begin{equation}
r = \frac{|p_{l_1} - p_{l_2}|_2}{|p_{r_1} -p_{r_2}|_2} - \frac{|p_{r_1} - p_{r_2}|_2}{|p_{l_1} -p_{l_2}|_2},
\end{equation}
where $p_{l_1}, p_{l_2}, p_{r_1}, p_{r_2}$ are the coordinates of left eye inner corner, left eye outer corner, right eye inner corner, and right eye outer corner respectively. With a threshold $\gamma$, we put the faces into the left or right bucket, when $r>\gamma$ or $r<-\gamma$. The other faces will be kept into the frontal bucket. We set $\gamma = 6$ in our experiments, as shown in Fig.~\ref{profile}. 

Based on the $N_{yaw}$ base anchors, to cover the roll variations, we rotate each anchor on the roll dimension. For example, we can get twenty-four templates by rotating the basic three anchors each $45^{\circ}$ from $0^{\circ}$ to $360^{\circ}$. Optionally, we can involve the pitch variations by directly projecting (rotating) along the pitch dimension. However, based on our experimental results, the anchors designed along the pitch view cannot further improve the performance but compromise the computational speed. Thus, in our final design, only anchors along the yaw and roll dimensions are utilized as shown in Fig.~\ref{templates}.

\begin{figure}[h]
    \centering
    \includegraphics[width=8.5cm]{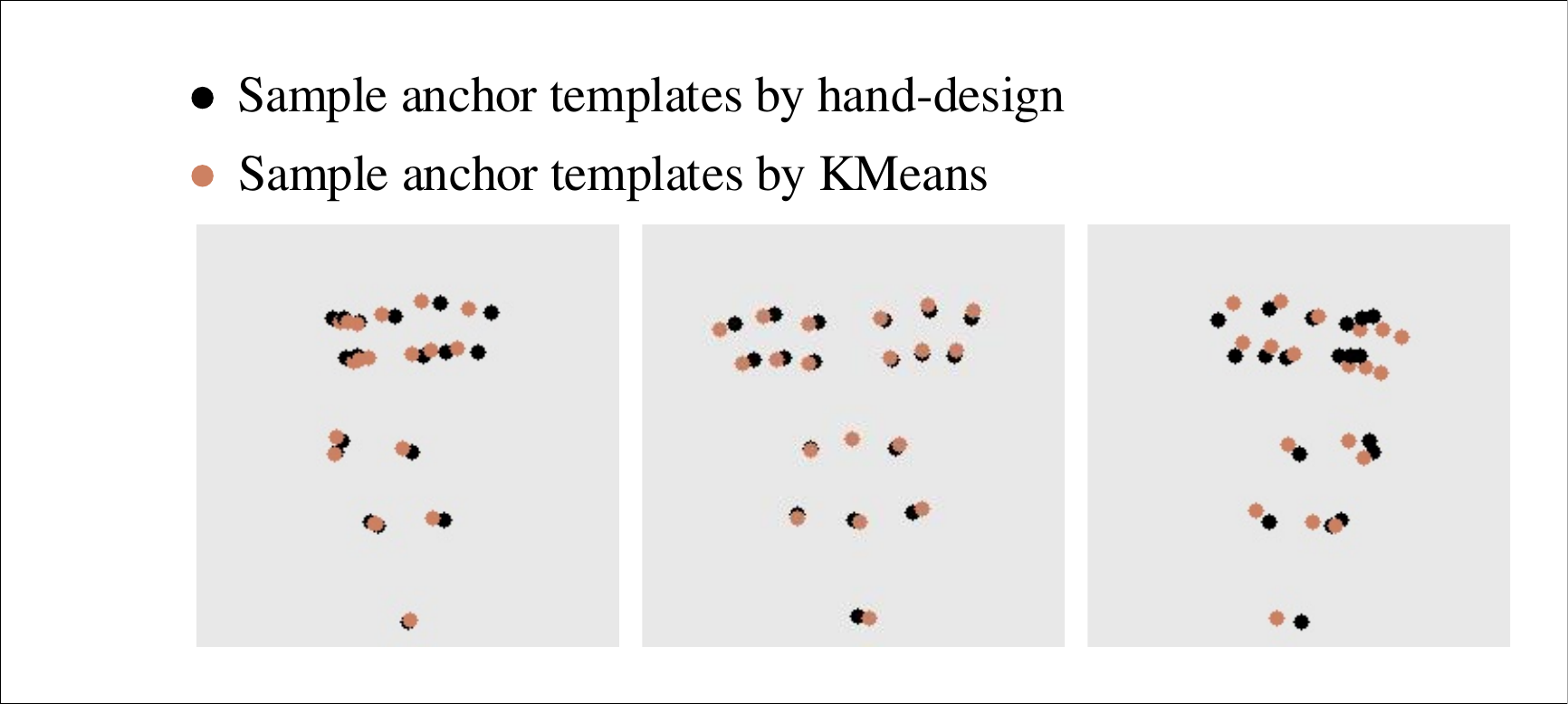}
    \caption{A comparison of three base anchors generated by hand-design approach and KMeans clustering based on AFLW~\cite{AFLW} dataset}
    \label{templates_kmeans}
\end{figure}

An alternative solution for the anchor design is based on the data distribution among the training faces. We first perform KMeans clustering, and we can generate a set of base anchors. One example is shown in Fig.~\ref{templates_kmeans}. We can see that the clustered anchors among all the training faces obtain similar anchors along the yaw direction, as discussed in hand-designed anchors. Following the similar steps for the hand-designed anchors, we can rotate the generated prototypes along the roll and pitch direction to generate more anchors. 

\subsubsection{Regression and Confidence Branch.}
Based on the anchor proposals, we design a new head structure which involves two branches: regression branch and confidence branch. Regression branch aims to regress the landmark coordinate offsets based on each anchor. Confidence branch assigns each anchor with a confidence score. Among all the anchor templates, those anchors which are close to the pose of the ground-truth face should be given higher confidence. 

As shown in Fig.~\ref{flow}, both the confidence branch and the regression branch are built upon the output feature map of the backbone network. While we set $h\cdot w$ anchors in the image, 
the output of the confidence and regression branch are $C_{con}\cdot h\cdot w$ and $C_{reg} \cdot h\cdot w$ respectively, where $C_{con}$ and $C_{reg}$ are denoted as the output channel number of the confidence branch and the regression branch respectively. Here $ C_{con}=K$ and $C_{reg}=K\cdot 2L$, 
where $K$, $L$ refer to the number of anchor templates on each anchor point and the number of facial landmarks respectively.

\begin{table}[htbp]
\begin{center}
    \scriptsize
    \begin{tabular}{cc}
    \toprule
    Symbol & Definition \\ 
    \midrule
    $A$ & A set of Anchor points on the spatial anchor grid \\
    $a$ & One anchor point $a\in A$ \\
    $T$ & A set of anchor templates as in Fig.~\ref{templates} \\
    $T(a,t)$ & Anchor template $t\in T$ centering at anchor point $a$ \\
    $T_j(a,t)$ & Landmark $j$ on anchor template $T(a,t)$ \\
    $O(a,t)$ & Output from the regression branch based on $T(a,t)$ \\
    $\overline{O}(a,t)$ & Ground-truth (GT) offsets based on $T(a,t)$ \\
    $C(a,t)$ & Output from the confidence branch based on $T(a,t)$ \\
    $\overline{C}(a,t)$ & Confidence GT label based on $T(a,t)$ \\
    \bottomrule
    \end{tabular}
    \caption{The definition of symbols}
    \label{symbol_definitions}
\end{center}
\end{table}

\subsection{Aggregate Step}

% On account of the confidence branch, each anchor face is assigned a confidence score, representing the reliability of its prediction. During split step, we split the search space with anchor templates by its spatial position and pose-level. Obviously, the real face is more close to some anchor templates and far away from others. All anchor templates will contribute to the final localization result, but with different weight.

Large-pose faces will increase the uncertainty for the landmark prediction. To address this problem, we propose to aggregate the predictions from different anchor templates. More specifically, we first set a threshold $C_{th}$ to pick up the reliable anchor predictions. The anchor predictions with low confidence scores are regarded as outliers and will be discarded. The remaining anchor predictions will be averaged by the weighted confidence for each prediction. As a result, the position of landmark $j$ can be obtained as the weighted average of the outputs of all anchor faces as:
% \begin{equation}
% \begin{aligned}
% \widetilde{S}_j = \frac{1}{\sum_{a,t}\widetilde{C}(a,t)} \sum_{a\in A, t\in T} \widetilde{C}_(a,t)\cdot (O_j(a,t) + T_j(a,t)),
% \end{aligned}
% \end{equation}
\begin{equation}
\begin{aligned}
\widetilde{S}_j = \frac{\sum_{a\in A, t\in T} \widetilde{C}(a,t)\cdot (O_j(a,t) + T_j(a,t))}{\sum_{a\in A, t\in T}\widetilde{C}(a,t)},
\end{aligned}
\end{equation}
where 
\begin{equation}
    \widetilde{C}(a,t)=\left\{
    \begin{aligned}
    0 &, \qquad  C(a,t) < C_{th}\\
    C(a,t) &, \qquad  others \\
    \end{aligned}
    \right.
\end{equation}
% \begin{equation}
% \end{equation}
The definition of the symbols can be found from Table~\ref{symbol_definitions}, and the threshold $C_{th}$ is set to $0.6$ in our experiments.

\subsection{Network Training}\label{confidence_label}

In this subsection, we will discuss the ground-truth setting for the regression and confidence branch as well as the related losses. For the regression branch, the target is to regress the offsets against each of the predefined anchor. The regression loss $L_{reg}$ can be defined as:
\begin{equation}
\begin{aligned}
L_{reg} = \sum_{a\in A, t\in T} & C(a,t) \sum_{j} |O_j(a,t) - \overline{O}_j(a,t)|,
\end{aligned}
\end{equation}
where $O_j(a,t)$ and $\overline{O}_j(a,t)$ refer to the prediction offsets and the ground-truth offsets for $j$th landmark. $C(a,t)$ is denoted as the confidence weight for the anchor template $T(a,t)$. The detailed symbol definitions can be found in Table~\ref{symbol_definitions}.

For the confidence branch, we set the targeted confidence output $\overline{C}(a,t)$ as the inverse L2 distance between the anchor pose $\bm{v_1}$ and the ground-truth pose $\bm{v_2}$ as  
$||\bm{v_1} - \bm{v_2}||_2$, where $\bm{v_1}, \bm{v_2}$ refer to flatten landmark coordinates. To normalize the pose difference, we perform a $\tanh$ operation as:
\begin{equation}
    \begin{aligned}
        \overline{C} = \tanh ((\frac{||\bm{v_1}- \bm{v_2}||_2}{\beta \cdot 2L})^{-1}),
        % \overline{C} = \tanh (||\bm{v_1}- \bm{v_2}||_2/ (\beta \cdot 2L),
    \end{aligned}
\end{equation}
where $\beta$ is a hyperparameter and $L$ refers to the count of facial landmarks.
The confidence loss is then defined as:
\begin{equation}
\begin{aligned}
L_{con} = & \sum_{a\in A, t\in T} (-C(a,t)\cdot \log \overline{C}(a,t)\\
&- (1-C(a,t))\cdot \log (1-\overline{C}(a,t))).
\end{aligned}
\end{equation}

The network is jointly supervised by the two loss functions above with end-to-end training.
The final training loss is then defined as:
\begin{equation}
    L_{total} = L_{reg} + \lambda \cdot L_{con}
\end{equation}
where $\lambda$ is a hyperparameter in our method, and it is insensitive to the localization accuracy in our experiments. %and the AnchorFace can be considered hyperparameter-free.%

\section{Experiment}
\subsection{Experiment settings}\label{sec:setting}
\subsubsection{Datasets.}
% \textbf{Datasets}. 
The experiments are evaluated on four challenging datasets, i.e. AFLW~\cite{AFLW}, 300W~\cite{300-W}, Menpo~\cite{Menpo2,Menpo-dataset}, and WFLW~\cite{LAB}. They are all widely used benchmarks in the facial landmark research area. More details about these datasets can be found in our supplementary materials.
\subsubsection{Evaluation metric.}
We adopt the normalized mean error (NME) for evaluation. The normalized mean error is defined as the average Euclidean distance between the predicted facial landmark locations $O_{i,j}$ and their corresponding ground-truth facial landmark annotations $\overline{O}_{i,j}$:
\begin{equation}
    NME = \frac{1}{N} \sum_{i=1}^{N} \frac{\frac{1}{L} \sum_{j=1}^{L} |O_{i,j}-\overline{O}_{i,j}|_2}{d}
\end{equation}
where $N$ is the number of images in the testing set, $L$ is the number of landmarks, and $d$ is the normalization factor. On AFLW dataset, we follow~\cite{setting} to use face size as the normalization factor. On Menpo dataset, we use the distance between left-top corner and right-bottom corner as the normalization factor. On 300W and WFLW dataset, we follow MDM~\cite{MDM} and~\cite{Menpo} to use the ``inter-ocular'' normalization factor, i.e. the distance between the outer eye corners.

In addition, on WFLW dataset, two further statistics~i.e. the area-under-the-curve (AUC)~\cite{AUC} and the failure rate~(which is defined as the proportion of failed detected faces) are measured for furthter analysis. Especially, any normalized error above $0.1$ is considered as a failure~\cite{LAB}.

\subsubsection{Implementation details.}
In our method, the original images are cropped and resized to a fixed resolution, i.e. $224\times 224$, according to the provided bounding boxes.
Anchor templates are generated based on KMeans clustering following~\ref{sec:anchor_template}, while anchor area and anchor grid are set as $56\times 56$ and $7\times 7$ respectively. 
Random rotation and translation are applied for data augmentation.
We apply the Adam optimizer with the weight decay of $1\times 10^{-5}$ and train the network for 50 epochs in total. The learning rate is set to $1\times 10^{-3}$ and divided by ten at 20-th, 30-th, 40-th epoch. $\beta=0.05$ and $\lambda=0.5$ are applied to all models across four benchmarks. If there are no special instructions, ShuffleNet-V2~\cite{ShuffleNet} is utilized as the backbone for our algorithm in this paper. 

\subsection{Comparison with the state-of-the-art methods}

For a fair comparison, we only compare the methods following the standard settings, as discussed in Section~\ref{sec:setting}. Therefore, those methods which are trained from external datasets or combined multiple datasets are not compared. Due to the space limit, comparisons about Menpo are reported in our supplementary materials.

\textbf{AFLW dataset}:  
We first evaluate our algorithm on the AFLW dataset.
The performance comparisons are given in Table~\ref{AFLW_experiment}. 
% and some localization examples are shown in Fig.~\ref{Samples}. 
It can be observed that, on this large dataset, our network outperforms the other approaches. 
As mentioned in Section~\ref{sec:setting}, AFLW contains lots of faces with large poses. Note that our method has a significant improvement on AFLW-Full set against AFLW-Frontal, which means that we achieve more robust localization performance on faces in unconstrained scenarios, including large pose. This essentially validates the superiority of our approach.

\begin{table}[!htbp]
\begin{center}
    \scriptsize
    \begin{tabular}{cccc}
    \toprule
    Methods & AFLW-Full & AFLW-Frontal \\ 
    \midrule
    LBF~\cite{LBF} & 4.24 & 2.74 \\
    CFSS~\cite{setting} & 3.92 & 2.69 \\
    CCL~\cite{CCL} & 2.72 & 2.17 \\
    TSR~\cite{300-w_setting}& 2.17 & - \\
    SAN~\cite{SAN} & 1.91 & 1.85  \\
    Wing~\cite{Wing} & 1.65 & - \\
    SA~\cite{SA} & 1.62 & - \\
    ODN~\cite{ODN} & 1.63 & 1.38 \\
    \midrule
    \textbf{AnchorFace} & \textbf{1.56} & \textbf{1.38} \\
    \bottomrule
    \end{tabular}
\end{center}
\caption{Normalized mean error~(\%) on AFLW-Full and AFLW-Frontal set.}
\label{AFLW_experiment}
\end{table}

\begin{table}[!htbp]
\begin{center}
    \scriptsize
    \begin{tabular}{cccc}
        \toprule
        Methods& Common& Challenge& Full\\ 
        \midrule
        MDM~\cite{MDM} & 4.83 & 10.14 & 5.88 \\
        Two-Stage~\cite{Two-stage} & 4.36 & 7.42 & 4.96 \\
        RDR~\cite{RDR} & 5.03 & 8.95 & 5.80 \\
        Pose-Invariant~\cite{3D3} & 5.43 & 9.88 & 6.30 \\
        SBR~\cite{SBR} & 3.28 & 7.58 & 4.10 \\
        PCD-CNN~\cite{PCD-CNN} & 3.67 & 7.62 & 4.44 \\
        LAB~\cite{LAB} & \textbf{2.98} & \textbf{5.19} & \textbf{3.49} \\
        SAN~\cite{SAN} & 3.34 & 6.60 & 3.98 \\
        ODN~\cite{ODN} & 3.56 & 6.67 & 4.17 \\ 
        \midrule
        \textbf{AnchorFace} & 3.12 & 6.19 & 3.72 \\
        \bottomrule
    \end{tabular}
\end{center}
\caption{Normalized mean error~(\%) on 300W Common subset, Challenging subset, and Full set.}
\label{300W_experiment}
\end{table}

% \begin{figure}[!htbp]
%     \centering
%     \includegraphics[width=10cm]{./figure/sample2.pdf}
%     \caption{Examples of the facial landmark localization results produced by our proposed method on the AFLW dataset.}
%     \label{Samples}
%     \vspace{-0.4cm}
% \end{figure}

\textbf{300W dataset}:
We compare our approach against several state-of-the-art methods on 300W Fullset. The results are shown in Table~\ref{300W_experiment}. 
% There are some researchers achieve high localization accuracy with cross dataset training, or introduce other input information, which is not fair to compare the results.
Since there are fewer large pose variations across the whole dataset and the cropped faces normally center near the image center point, 300W dataset is not very challenging compared with the other three benchmarks. However, our algorithm still can achieve promising localization performance with an efficient speed at $\sim$45 fps with batch size 1. Compared with LAB~\cite{LAB}, which is slightly better than our method, our approach is much faster (45 fps vs 17 fps).

\begin{table}[!htbp]
\begin{center}
    \tiny
    \begin{tabular}{ccccc}
        \toprule
        Methods& NME(\%)& Failure Rate(\%)& AUC & Flops(G)\\ 
        \midrule
        CFSS~\cite{setting}  & 9.07 & 29.40 & 0.3659 & - \\
        DVLN~\cite{DVLN} & 6.08 & 10.84 & 0.4551 & - \\
        LAB~\cite{LAB}  & 5.27 & 7.56 & 0.5323 & 18.85 \\
        SAN~\cite{SAN} & 5.22 & 6.32 & 0.5355 & - \\
        Wing~\cite{Wing} & 5.11 & 6.00 & 0.5504 & 5.40\\
        AVS~\cite{AVS} & 5.25 & 7.44 & 0.5034 & - \\
        AWing~\cite{AWING} & 4.36 & \textbf{2.84} & 0.5719 & 26.79 \\ 
        HRNET~\cite{HRNET} & 4.60 & 4.64 & 0.5237 & - \\
        \midrule
        \textbf{AnchorFace} & 4.62 & 4.20 & 0.5516  & \textbf{1.71}\\
        \textbf{AnchorFace*} & \textbf{4.32} & 2.96 & \textbf{0.5769} & 5.30\\
        \bottomrule
    \end{tabular}
\end{center}
\caption{Evaluation results about WFLW test set for NME(\%), Failure Rate@0.1(\%) and AUC@0.1. \textbf{*} means we adopt HRNet-18~\cite{HRNET} as the backbone network to implement AnchorFace.}
\label{WFLW_experiments}
\end{table}

\textbf{WFLW dataset}:
A comparison of the performance from our proposed approach as well as state-of-the-art methods on WFLW dataset is shown in Table~\ref{WFLW_experiments}. Considering the difficulty of WFLW, we attempt to adopt HRNet-18~\cite{HRNET} as the backbone to implement our AnchorFace. As indicated in Table~\ref{WFLW_experiments}, AnchorFace equipped with HRNet-18~\cite{HRNET} achieves the best performance on NME \& AUC metrics and achieves the second best performance on Failure Rate metric. Furthermore, the efficent AnchorFace with ShuffleNet-V2~\cite{ShuffleNet} also achieves impressive results considering the huge computation cost of other methods. This also directly verifies the versatility of our method. Detailed results about each subset of WFLW are reported in our supplementary materials.

\subsection{Model analysis}
Our proposed AnchorFace introduces a novel split-and-aggregate strategy based on anchor design to address the face alignment across large poses. In this section, we perform further analysis of its mechanism.

\begin{figure}[htbp]
    \centering
    \includegraphics[width=8cm]{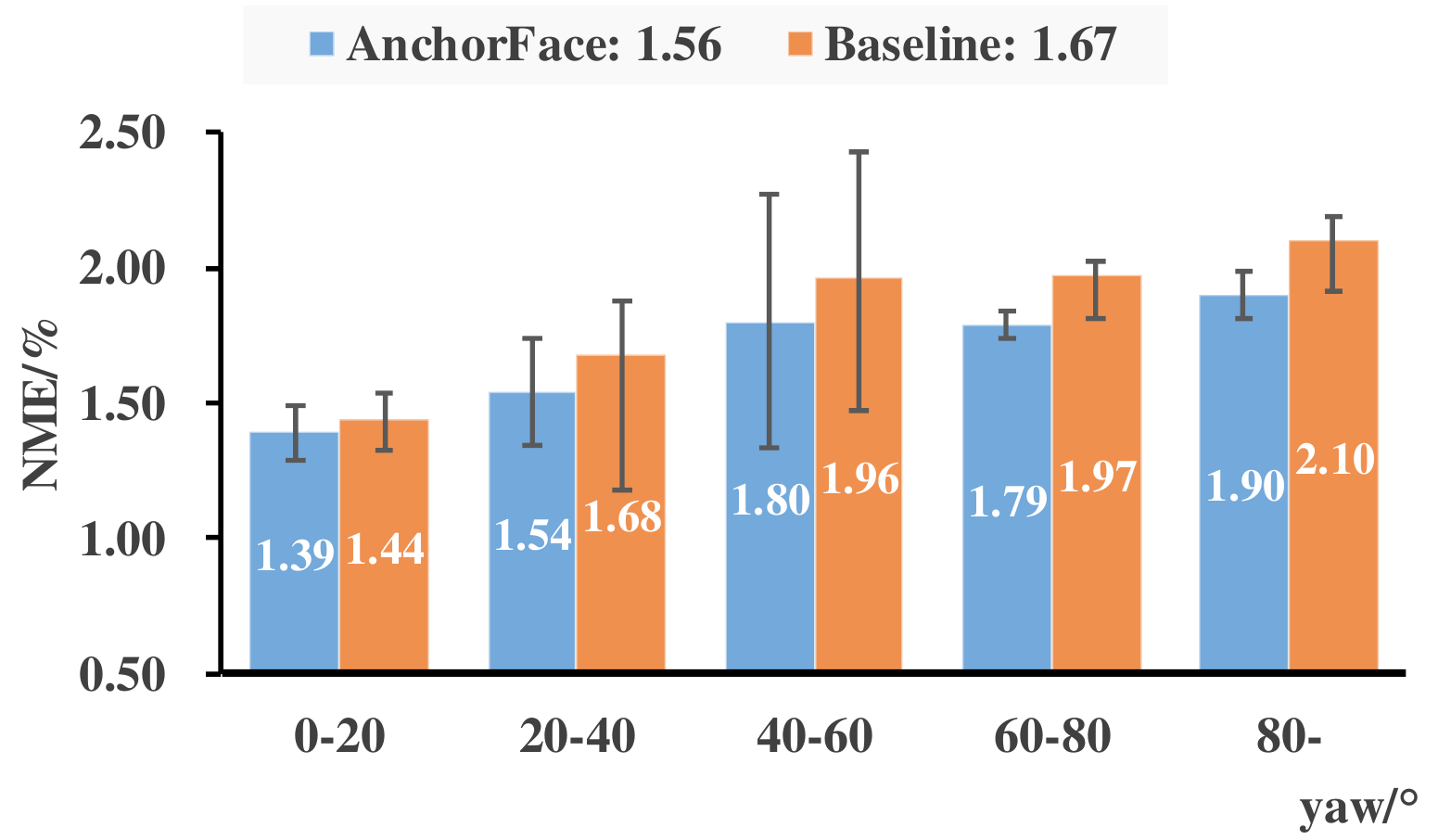}
    \caption{A comparison of baseline and AnchorFace across yaw dimension}
    \label{reason}
\end{figure}

\textbf{Anchor design}. Anchor templates serve as regression references to split the search space in our proposed approach. In comparison with directly regressing target landmark coordinates in whole 2D space, regress offsets based on anchor templates can simplify the search space and boost the robustness of localization accuracy. We conduct several experiments on AFLW dataset and make statistics across yaw dimension which is shown in Fig.~\ref{reason}. It is quite clear that AnchorFace significantly outperforms the baseline with lower NME and smaller variances in each subinterval especially for large pose, which can well verify our assumptions.

\textbf{Split-and-aggregate strategy}. In our proposed algorithm, we follow the divide-and-conquer way to address the challenges for face alignment across large poses. To verify its effectiveness, we adopt \emph{Pearson correlation coefficient} to measure the correlation between \textbf{confidence scores $C(a, t)$} and \textbf{prediction errors $|O(a, t) - \overline{O}(a,t)|_2$}:
\begin{equation}
    \begin{split}
         \emph{P} &= \frac{1}{N} \sum_{i=1}^{N} [\boldsymbol{r}_{a, t}^{i}(|O(a, t) - \overline{O}(a,t)|_2, C(a, t))]
    \end{split}
\end{equation}
Where $\boldsymbol{r}$ represents the calculation function of \emph{Pearson correlation coefficient}. We conduct experiments on AFLW dataset and get P = -0.82, which means a strong negative correlation between them. In other words, anchor template with larger confidence score can achieve more accurate predicted landmarks. It can help filter prediction outliers and aggregate remaining predictions to mitigate the uncertainty of the localization result on a single anchor face. Due to the confidence score is defined as mathematical modeling of the distance between the anchor pose and the ground-truth pose, we can come to another conclusion that \emph{closer} anchors tend to achieve more accurate localization, which also directly proves our search space split strategy based on anchors. Comparison details can be found in our ablation studies and we show several intuitive samples in our supplementary materials.

% \begin{figure}
%     \centering
%     \includegraphics[width=11cm]{figure/Analysis.pdf}
%     \caption{Mechanism verification: 1. Anchor points set on sample image; 2. One template sampled from blue anchor point; 3. One template sampled from green anchor point; 4. Comparison of the predicted result of above two templates, and ground-truth is in red}
%     \label{analysis}
%     \vspace{-0.4cm}
% \end{figure}
\iffalse
\begin{figure}
    \centering
    \includegraphics[width=8cm]{figure/Analysis.pdf}
    \caption{Algorithm analysis based on different anchors. First column shows anchor grid settings, the green and blue anchors are two selected samples. Second and third column shows two anchor templates with prediction scores, which are randomly selected from the previous two anchors. The last column shows the final prediction results while the groud-truth is in red}
    \label{analysis}
    % \vspace{-0.4cm}
\end{figure}
\fi
\subsection{Ablation study}\label{sec:ablation}
% Our framework is divided into two steps: Split and Aggregate. 
In this section, we perform the ablation study for our proposed algorithm on the AFLW dataset, which is a challenging benchmark with large pose variations. More specifically, we divide the test set into four subsets according to the yaw dimension, i.e. Light ($0^{\circ}\sim 30^{\circ}$), Medium ($30^{\circ}\sim 60^{\circ}$), Large ($60^{\circ}\sim 90^{\circ}$), and Heavy ($90^{\circ}\sim $). Normalized mean error is utilized to evaluate the performance of our algorithm. Without explicitly specified, we use anchor templates as KMeans-24 (KMeans clustering to generate 24 anchor templates), anchor area as $56 \times 56$, anchor grid as $7 \times 7$, and the aggregating strategy is weighted average for ablation. 

\subsubsection{Comparison with the regression baseline.}
Table~\ref{ABLA_DIRECT} compares the performance of our proposed approach with the baseline of direct regression on AFLW dataset. 
``Baseline'' directly maps the discriminative features to the target landmark coordinates with ShuffleNet-V2 backbone. 
A fully connected layer with length $2L$ is used as the output of the baseline network.
As shown in Table ~\ref{ABLA_DIRECT}, our proposed anchor-based method significantly outperforms the baseline by a large margin across yaw variations.The improvements are attributed to two reasons. First, the anchor design can significantly reduce the search space and simplify the regression problem. Second, the aggregating of different anchors can further improve model robustness. 

%  \begin{table}[!htbp]
%  \caption{A comparison of direct regression and anchor-based regression.}
%  \begin{center}
%      \footnotesize
%      \begin{tabular}{cccccc}
%      \toprule
%      Method & Full & Light & Medium & Large & Heavy \\
%      \midrule
%      Baseline & 1.67 & 1.46 & 1.92 & 1.99 & 2.13 \\ 
%      \textbf{AnchorFace} & \textbf{1.56} & \textbf{1.40} & \textbf{1.74} & \textbf{1.80} & \textbf{1.96}\\
%      \bottomrule
%      \end{tabular}
%  \end{center}
%  \label{ABLA_DIRECT}
%  \vspace{-1cm}
%  \end{table}

\subsubsection{Comparison of various split configurations.}
Due to the challenges from the large-pose faces, we propose a set of anchor templates as references for regression to split search space. Split strategy consists of three hyper-parameters: anchor templates, anchor area, and anchor grid, as shown in Fig.~\ref{templates}. 

\begin{table}[!htbp]
\begin{center}
    \scriptsize
    %\resizebox{\textwidth}{!}{
    \begin{tabular}{cccccccccccc}
    \toprule
    % Method & AFLW-Full & $0^{\circ}\sim 30^{\circ}$ & $30^{\circ}\sim 60^{\circ} $ & $60^{\circ}\sim 90^{\circ}$ & $90^{\circ}\sim $ \\
    Method & Full & Light & Medium & Large & Heavy \\
    \midrule
    Baseline & 1.67 & 1.46 & 1.92 & 1.99 & 2.13 \\ 
    \textbf{AnchorFace} & \textbf{1.56} & \textbf{1.40} & \textbf{1.74} & \textbf{1.80} & \textbf{1.96}\\
    \bottomrule
    \end{tabular}
\end{center}
\setlength{\abovecaptionskip}{0pt}
\caption{A comparison of direct regression and anchor-based regression.}
\label{ABLA_DIRECT}
\vspace{-0.5cm}
\end{table}

\begin{table}[!htbp]
    \begin{center}
        \scriptsize
        \begin{tabular}{cccccc}
        \toprule
            % Anchor area & Full & $0^{\circ} \sim 30^{\circ}$ & $30^{\circ} \sim 60^{\circ}$ & $60^{\circ} \sim 90^{\circ}$ & $90^{\circ} \sim $ \\
            Template & Full & Light & Medium & Large & Heavy \\
        \midrule
            Kmeans-3 & 1.60 & 1.43 & 1.80 & 1.83 & 2.10 \\
            Kmeans-24 & \textbf{1.56} & \textbf{1.40} & \textbf{1.74} & \textbf{1.80} & \textbf{1.96} \\
            Kmeans-48 & 1.58 & 1.41 & 1.79 & 1.82 & 2.06 \\
            HandDesign-24 & 1.58 & 1.42 & 1.76 & 1.84 & 2.00 \\
        \bottomrule
        \end{tabular} 
    \end{center}
    \setlength{\abovecaptionskip}{0pt}
    \caption{A comparison of different template settings.}
    \label{template-count}
\end{table}

\textbf{Anchor template} plays a voting role in our method as a reference for regression. As mentioned in Section~\ref{sec:anchor_template}, we get three basic template faces from the training dataset by hand design or KMeans clustering. Then we do some transformations to get more templates, corresponding to the pose variations in yaw, roll, and pitch dimension. By comparing KMeans-24 against HandDesign-24 in Table~\ref{template-count}, KMeans is better than hand-design approach based on the same anchor number (24). The potential reason is that KMeans utilizes more data features to generate the base anchors, which should be more general compared with hand-designed based anchors. Besides, as shown in Table~\ref{template-count}, 24 may be a good option for the number of anchor templates compared with 3 or 48 in our algorithm.

% \begin{table}[!htbp]
%     \setcaptionmargin{0.2cm}
%     \caption{A comparison of direct regression and anchor-based regression.}
%     \begin{center}
%     \footnotesize
%     \begin{tabular}{cccccc}
%         \toprule
%         Method & Full & Light & Medium & Large & Heavy \\
%         \midrule
%         Baseline & 1.67 & 1.46 & 1.92 & 1.99 & 2.13 \\ 
%         \textbf{AnchorFace} & \textbf{1.56} & \textbf{1.40} & \textbf{1.74} & \textbf{1.80} & \textbf{1.96}\\
%         \bottomrule
%     \end{tabular}
%     \end{center}
%     \label{ABLA_DIRECT}
% \vspace{-0.4cm}
% \end{table}  

\textbf{Anchor area} is the area where we set anchors in the image for the spatial domain. As the input image is cropped and resized to $224\times 224$ and the face is around the image center, we set anchor points at a center area with size $14\times 14$, $28\times 28$, $56\times 56$, $128 \times 128$, respectively. As shown in Table~\ref{area}, $56\times 56$ around the image center would be a good choice for putting anchors in the spatial domain. 

\begin{table}[!htbp]
    \begin{center}
        \scriptsize
        \begin{tabular}{cccccc}
        \toprule
            % Anchor area & Full & $0^{\circ} \sim 30^{\circ}$ & $30^{\circ} \sim 60^{\circ}$ & $60^{\circ} \sim 90^{\circ}$ & $90^{\circ} \sim $ \\
            Anchor area & Full & Light & Medium & Large & Heavy \\
        \midrule
            $112\times 112$ & 1.58 & 1.40 & 1.79 & 1.83 & 2.06 \\
            $56\times 56$ & \textbf{1.56} & \textbf{1.40} & \textbf{1.74} & \textbf{1.80} & \textbf{1.96} \\
            $28\times 28$ & 1.58 & 1.42 & 1.79 & 1.82 & 2.04 \\
            $14\times 14$ & 1.58 & 1.41 & 1.81 & 1.83 & 2.01 \\
        \bottomrule
        \end{tabular}
    \end{center}
    \setlength{\abovecaptionskip}{0pt}
    \caption{A comparison of different anchor area settings.}
    \label{area}
\end{table}

\textbf{Anchor grid} defines how many anchors we set in the anchor area. For example, $7\times7$ means we sample 49 spatial points in a $7 \times 7$ grid from the anchor area to generate the anchor templates. As shown in Table~\ref{count}, it is a good choice to set at $7\times7$.

\begin{table}[!htbp]
    \begin{center}
        \scriptsize
        \begin{tabular}{cccccc}
        \toprule
            % Anchor count & Full & $0^{\circ} \sim 30^{\circ}$ & $30^{\circ} \sim 60^{\circ}$ & $60^{\circ} \sim 90^{\circ}$ & $90^{\circ} \sim $ \\
            Anchor count & Full & Light & Medium & Large & Heavy \\
        \midrule
            $3\times 3$ & 1.58 & 1.40 & 1.78 & 1.82 & 2.17 \\
            $5\times 5$ & 1.57 & 1.40 & 1.77 & 1.82 & 2.05 \\
            $7\times 7$ & \textbf{1.56} & \textbf{1.40} & \textbf{1.74} & \textbf{1.80} & \textbf{1.96}\\
            $13\times 13$ & 1.56 & 1.40 & 1.75 & 1.81 & 2.07 \\
        \bottomrule
        \end{tabular}
    \end{center}
    \setlength{\abovecaptionskip}{0pt}
    \caption{A comparison of different anchor grid settings.}
    \label{count}
    \vspace{-0.4cm}
\end{table}

\subsubsection{Comparison of various aggregate strategies.}\label{aggregate}
To mitigate the uncertainty of the localization result on a single anchor face, we aggregate the predictions from different anchor templates. We introduce three aggregate strategies: Mean, Argmax, confidence weighted voting (Weighted). As shown in Table~\ref{aggregation}, aggregating the predictions with weighted confidences can obtain superior results compared with the argmax choice without aggregating. Besides, the confidence generated by the confidence branch is important if we compare the strategy of ``Weighted'' and ``Mean''.
\begin{table}[!htbp]
    \begin{center}
        \scriptsize
        \begin{tabular}{cccccc}
        \toprule
            % Aggregate & Full & $0^{\circ} \sim 30^{\circ}$ & $30^{\circ} \sim 60^{\circ}$ & $60^{\circ} \sim 90^{\circ}$ & $90^{\circ} \sim $ \\
            Aggregate & Full & Light & Medium & Large & Heavy \\
        \midrule
            Argmax & 1.58 & 1.42 & 1.76 & 1.83 & 2.00 \\
            Weighted & \textbf{1.56} & \textbf{1.40} & \textbf{1.74} & \textbf{1.80} & \textbf{1.96} \\
            Mean  & 1.61 & 1.43 & 1.80 & 1.89 & 2.22 \\
        \bottomrule
        \end{tabular}
    \end{center}
    \setlength{\abovecaptionskip}{0pt}
    \caption{A comparison of different aggregation strategies.}
    \label{aggregation}
\end{table}
% \begin{table}[htbp]
%     \centering
%     \caption{NME based on different anchors}
%     \label{tab:my_label}
%     \begin{tabular}{ccccc}
%     \toprule
%     Anchors & $<0.2$ & $0.2-0.4$ & $0.4-0.6$ & $>0.6$ \\
%     \midrule
%     NME(\%) & 1.66 & 1.60 & 1.57 & 1.56 \\
%     \bottomrule
%     \end{tabular}
% \vspace{-1cm}
% \end{table}

\section{Conclusions}
In this paper, a novel split-and-aggregate strategy is proposed for large-pose faces. By introducing an anchor-based design, our proposed approach can simplify the regression problem by splitting the search space. Moreover, aggregating the prediction results contributes to reducing uncertainty and improving the localization performance. As validated on four challenging benchmarks, our proposed AnchorFace obtains state-of-the-art results with extremely fast inference speed.
\newpage
\bibliography{anchorface}
\end{document}